\documentclass[10pt,twocolumn]{iopart}

\usepackage[pdftex]{graphicx}
\usepackage{iopams}
\usepackage{caption}
\usepackage{ulem}
\usepackage{xcolor}
\usepackage{dblfloatfix}
\expandafter\let\csname equation*\endcsname\relax

\expandafter\let\csname endequation*\endcsname\relax

\usepackage{amsmath}

\begin{document}

\title[Tuning the activation function to optimize the predictions of a RC]{Tuning the activation function to optimize the forecast horizon of a reservoir computer}

\author{L A Hurley$^1$, J G Restrepo$^2$ and S E Shaheen$^{1,3,4}$}

\address{$^1$Department of Electrical, Computer and Energy Engineering, University of Colorado at Boulder}
\address{$^2$Department of Applied Mathematics, University of Colorado at Boulder}
\address{$^3$Renewable and Sustainable Energy Institute (RASEI)}
\address{$^4$Department of Physics, University of Colorado at Boulder}
\ead{lauren.hurley@colorado.edu}
\vspace{10pt}
\begin{indented}
\item[]December 2023
\end{indented}



\begin{abstract}
Reservoir computing is a machine learning framework where the readouts from a nonlinear system (the {\it reservoir}) are trained so that the output from the reservoir, when forced with an input signal, reproduces a desired output signal. A common implementation of reservoir computers is to use a recurrent neural network as the reservoir. The design of this network can have significant effects on the performance of the reservoir computer. In this paper we study the effect of the node activation function on the ability of reservoir computers to learn and predict chaotic time series. We find that the {\it Forecast Horizon} (FH), the time during which the reservoir's predictions remain accurate, can vary by an order of magnitude across a set of 16 activation functions used in machine learning. By using different functions from this set, and by modifying their parameters, we explore whether the entropy of node activation levels or the curvature of the activation functions determine the predictive ability of the reservoirs. We find that the FH is low when the activation function is used in a region where it has low curvature, and a positive correlation between curvature and FH. For the activation functions studied we find that the largest FH generally occurs at intermediate levels of the entropy of node activation levels. Our results show that the performance of reservoir computers is very sensitive to the activation function shape. Therefore, modifying this shape in hyperparameter optimization algorithms can lead to improvements in reservoir computer performance.

\end{abstract}


%
\vspace{2pc}
\ioptwocol

\section{Introduction}
Reservoir computing is a type of recurrent neural network (RNN) with high performance in learning and predicting the dynamics of chaotic systems. Its use is growing rapidly thanks to its simplified training process \cite{guiding, modelf, noise, variability} and versatile physical implementations \cite{tanaka2019recent}. A Reservoir Computer (RC) consists of an input layer, a high-dimensional nonlinear dynamical system (the reservoir), and a trainable output layer. The advantage of this framework is that only the output layer needs to be trained, while the reservoir remains fixed. A common kind of RC is an Echo State Network (ESN), where the reservoir is a recurrent network of randomly connected nodes with fixed weights \cite{datad, overview, diverse}. Reservoir computers with this implementation can achieve high accuracy in pattern analysis and prediction tasks \cite{ESN, rain, systematic}.  

\begin{figure*}[!t]
  \begin{center}
  \includegraphics[width=0.7\textwidth]{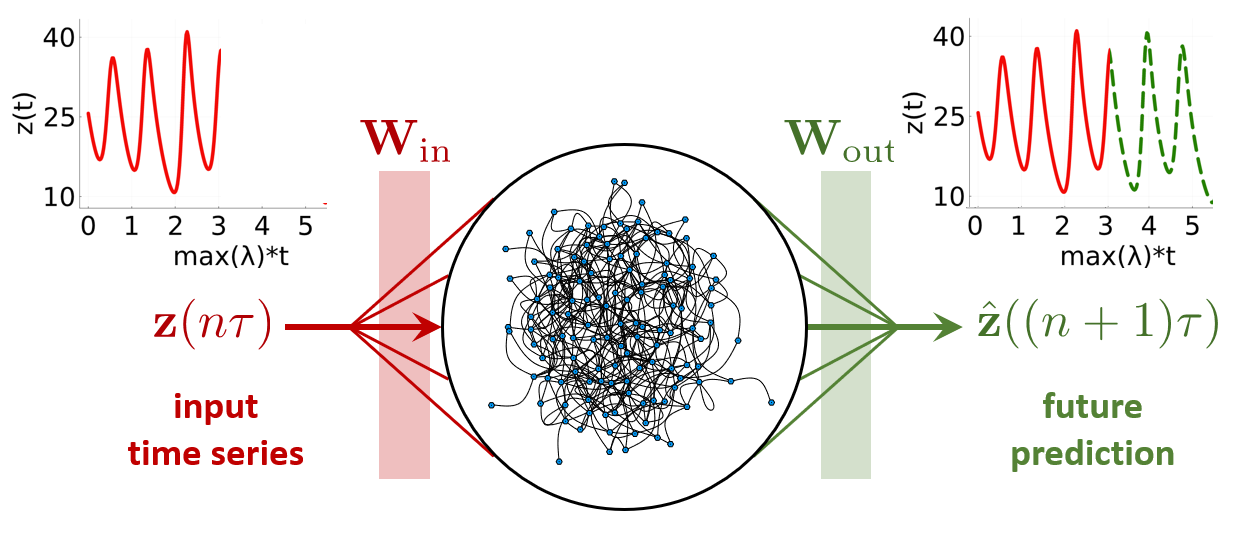}\\
  \caption{Diagram of a reservoir computer. The input time series and future prediction graphs shown are the z-coordinate of the Lorenz system. The reservoir in the image is a representation of a randomly generated 125-node reservoir.}\label{RC}
  \end{center}
\end{figure*}
The implementation of an ESN reservoir computer requires various parameter and structural choices, including the size of the reservoir, the structure of the reservoir network, the node activation function, and the process by which the output layer is trained. These choices can have a significant effect on the performance of the reservoir computer, and therefore have received much attention. Ref.~\cite{systematic} studies systematically the effect of reservoir size, spinup time, amount of training data, normalization, noise, and input bias on RC performance, while Ref.~\cite{carroll} studies the effect of reservoir network structure. Hyperparameter optimization techniques \cite{bergstra} are often used to select many of these parameters. An important feature of an ESN reservoir computer is the activation function used for the dynamics of each node in the reservoir. For other types of machine learning architectures there have been targeted efforts to optimize the activation function (e.g., see \cite{impact,searching, benchmark, mish,pflu,logish}), some including the use of genetic and automated searching algorithms \cite{searching, genetic}. In the context of ESN reservoir computers, with a few exceptions, the choice of activation function has remained relatively unexplored. Refs.~\cite{thesis,breaking_sym} studied how the hyperbolic tangent activation function can be modified to remove the negative effects of symmetries in the equations. Refs.~\cite{overview, xinli,shapes} compare the performance of various activation functions, and Ref.~\cite{imp_ESN} shows how a combination of the usual hyperbolic tangent activation function with its derivatives can lead to higher performance. In this paper, we investigate more systematically how the performance of the reservoir depends on the activation function. Our metric of performance is the Forecast Horizon (FH), which measures how long the RC can accurately predict the future evolution of a system (see Section 2). We calculate the FH for a set of 16 commonly used or recently proposed activation functions, and find that there is a large variability in the resulting FH, with FHs varying by an order of magnitude. In addition, we explore how the FH changes as some of the parameters in these functions are changed. We explore numerically how the FH depends on the curvature and monotonicity of the activation function and on the entropy of reservoir states, features which have previously been hypothesized to be correlated with reservoir performance \cite{ase1,entropy1, carroll_entropy1,carroll_entropy2}.

Our paper is organized as follows. In Sec.~\ref{background} we review the operational details of reservoir computers and define our performance metric, the FH. In Sec.~\ref{impact} we present our results on how the activation function impacts the FH. In Sec.~\ref{metrics} we explore how the FH depends on various metrics related to the dynamics of the operation of the reservoir. Finally, in Sec.~\ref{conclusion} we discuss our results.

\section{Forecasting with Reservoir Computers}\label{background}

In this paper we are interested in the performance of ESN reservoir computers when predicting the evolution of a chaotic system. 
Suppose that it is desired to train the reservoir computer to predict the evolution of such a system based on an existing time series of the system's coordinates ${\bf z}$ sampled every $\tau$ time units during a training time $T$, $\{{\bf z}(n\tau)\}_{n=1}^{T/\tau}$. This time series is used to drive the internal states of the reservoir, ${\bf r}$, by using a fixed input matrix $W_{\text{in}}$ as
\begin{align}
{\bf r}((n+1)\tau) = f(A {\bf r}(n\tau) + W_{\text{in}} {\bf z}(n\tau)),\label{ridge}
\end{align}
where the matrix $A$ represents the internal connectivity of the reservoir network and $f(\cdot)$ is the activation function, which is assumed to act on each component of its vector input and is the focus of this paper. The reservoir output is given by $\hat {\bf z}((n+1)\tau) = W_{\text{out}} {\bf r}((n+1)\tau)$, where the output matrix is chosen so that the output predicts as closely as possible the future state of the reservoir, i.e., $\textbf{z}((n+1)\tau)\approx \hat{\textbf{z}}((n+1)\tau)$.  This can be done by a single least-squares regression that minimizes the cost function
\begin{align}
\sum_{n = 1}^{T/\tau}\| \hat {\bf z}(n\tau) - {\bf z}(n\tau)\|^2 + \lambda \text{Tr}(W_{\text{out}}  W_{\text{out}} ^T),
\end{align}
where the last term, with $\lambda \geq 0$, is a regularization term to prevent overfitting.
The activation function $f$ must be nonlinear so that the RC can learn nonlinear relationships between inputs and outputs \cite{imp_ESN, survey}. A common choice for the activation function is a hyperbolic tangent, $f(x) = \tanh(x)$ \cite{Ott}, but other choices are possible \cite{shapes} and will be explored here. 
\begin{figure*}[!b]
  \begin{center}
  \includegraphics[width=0.8\linewidth]{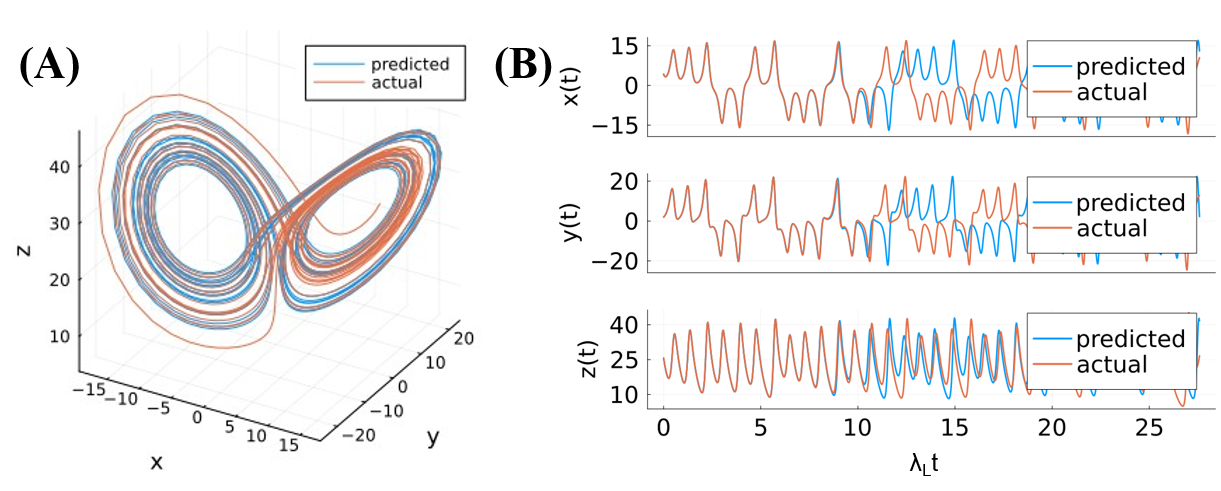}\\
  \caption{(A) Plot of the actual (red) and predicted (blue) Lorenz system coordinates (see text for reservoir parameters). The activation function used was the swish function with parameter $\beta = 0.6$ [see Equation~\ref{swish}]. (B) Plot of the actual (red) and predicted (blue) Lorenz coordinates as a function of Lyapunov time.}\label{butterfly}
  \end{center}
\end{figure*}
Reservoir computers can be used for recognition or prediction tasks \cite{systematic, Ott, recognition, prediction}. For prediction, a commonly used benchmark is the chaotic Lorenz System \cite{Lorenz}:
\begin{equation}\label{lorenz}
\begin{array}{l}
    \frac{dx}{dt} = \sigma(y-x),\\
    \frac{dy}{dt} = x(\rho -z) - y,\\
    \frac{dz}{dt} = xy - \beta z,
\end{array}
\end{equation}
with $\sigma = 10$, $\rho = 28$, and $\beta = 8/3$. In order to test the prediction performance of a given activation function $f$, we first train a RC as described above using a time series $\{{\bf z}(n\tau)\}_{n=1}^{T/\tau} = \{x(n\tau),y(n\tau),z(n\tau)\}_{n=1}^{T/\tau}$ obtained from numerical solution of Equation~(\ref{lorenz}). Once the reservoir is trained, we switch to prediction mode for $n\tau > T$ by using the output of the reservoir at time $n\tau$, $\hat z((n+1)\tau)$, as its input for the next prediction step, and so on. If the training is successful, the resulting autonomous dynamics reproduce those of the original Lorenz system, including features such as Lyapunov exponents \cite{Ott}. However, since the dynamics are chaotic, the reservoir can only track the dynamics of the original Lorenz system for a limited time. This is illustrated in Figure \ref{butterfly}(A), which shows the generated data for a Lorenz system (red) and the results generated by an autonomous RC in prediction mode (blue). To better visualize the results, Figure \ref{butterfly}(B) shows the \textit{x-}, \textit{y-}, and \textit{z-}coordinates of this system on separate plots. The horizontal axis shows time in prediction mode in Lyapunov times, i.e., the maximum Lyapunov exponent of the Lorenz system, $\lambda_{{\text L}}\approx 0.906$, multiplied by time. Here, it is visible that the prediction and actual curves overlap until about 10 Lyapunov time units. 
\begin{figure*}[!b]
  \begin{center}
  \includegraphics[width=0.9\textwidth]{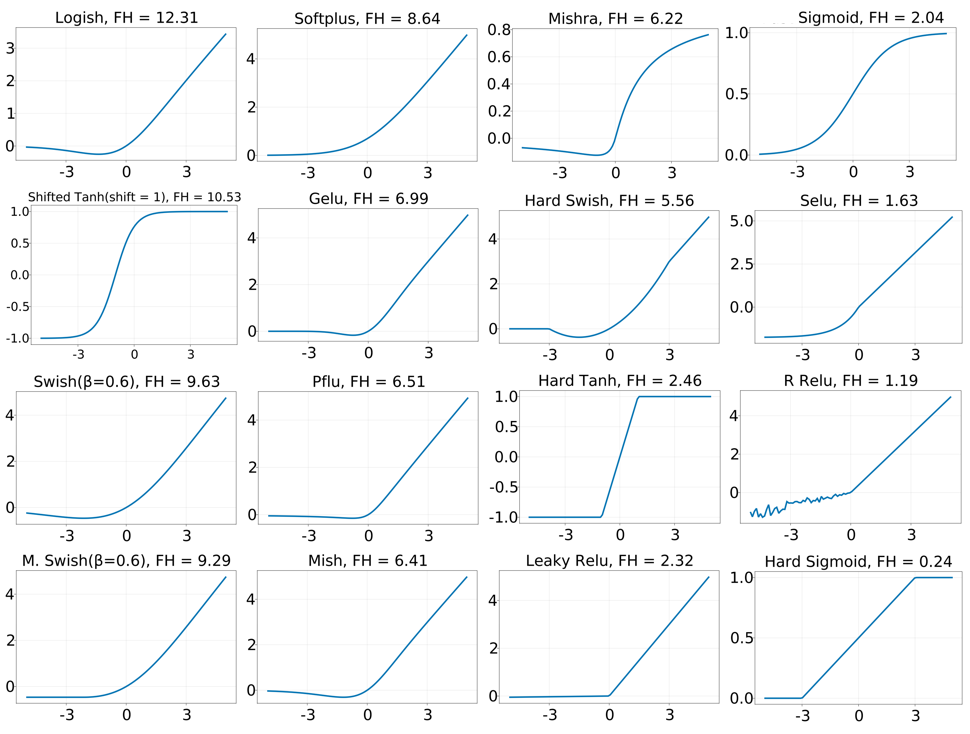}\\
  \caption{Graphs of the activation functions studied in this paper and their associated forecast horizons. The FHs shown are an average over 50 trials with reservoirs of size $N = 300$. The definitions of the activation functions are in \ref{appendixA}.}\label{graphs}
  \end{center}
\end{figure*}

In this work, we define the FH as the time, measured in Lyapunov times, when the predicted $x$ coordinate first deviates from the actual $x$ coordinate by a value of 5, i.e.,
\begin{align}
\text{FH} = \lambda_{{\text L}} \inf\{ t > 0, |x(t) - \hat x(t) | > 5  \}.
\end{align}
As in \cite{breaking_sym}, the threshold is chosen to be approximately 15\% of the total range for a given coordinate.  For the example shown in Figure \ref{butterfly}(B) the FH is approximately $10.5$, which corresponds visually to the part where actual and predicted coordinates (red and blue, respectively) start to diverge. We note that the specific choice of threshold, and the choice of using the $x$ coordinate only in the definition, do not affect our results: once the predicted values of the coordinates start to appreciably differ from the actual values, the error grows very quickly in all coordinates. Thus, other choices would yield slightly different FH values, but the qualitative dependence of the FH on reservoir parameters and activation function would be preserved.

We used the Julia package ReservoirComputing.jl to simulate the ESN reservoirs \cite{julia}. Unless otherwise noted, the parameters used were as follows: the reservoir size (number of nodes) was $N = 300$. The reservoir matrix $A$ was constructed by choosing each entry to be $1$ with probability $6/N$ and $0$ otherwise, and then rescaling the matrix so that its spectral radius was $1.2$. The input matrix W$_\text{in}$ was chosen by assigning a random number uniformly distributed in $[-0.1,0.1]$ to each entry. The training method is a standard Ridge Regression with a regularization parameter $\lambda = 0$ [See Equation 2], the training time $T$ was 5000, the prediction time was 1250, and the sampling time $\tau$ was 0.02. For the example shown in Figure 2, the activation function used was the {\it swish} function, $f(x) = x/(1+\exp(-\beta x))$ with $\beta = 0.6$. 

\section{Impact of activation function on FH}\label{impact}

\subsection{Comparing different activation functions}
\begin{table*}
\centering
\begin{tabular}{@{}lcclcc}
\br
Non-monotonic & FH & Standard Error & Monotonic Function & FH & Standard Error\\
Function & & & & & \\
\mr
 Logish \cite{logish} & 12.31 & 0.268 & Shifted tanh (bias = 1) & 10.53 & 0.207\\ 
 Swish ($\beta=0.6$) & 9.63 & 0.321 & Monotonic swish ($\beta=0.6$) & 9.29 & 0.321 \\
 Gelu & 6.99 & 0.359 &  Softplus & 8.64& 0.342 \\
 Pflu \cite{pflu} & 6.51 & 0.410 & Hard tanh & 2.46 & 0.236 \\
 Mish \cite{mish} & 6.41 & 0.317 & Leaky Relu & 2.32 & 0.190\\
 Mishra \cite{benchmark} & 6.22 & 0.354 & Sigmoid & 2.04 & 0.605\\
 Hard swish & 5.56 & 0.211 & Selu & 1.63 & 0.093\\ 
 & & & R Relu & 1.19 & 0.094 \\
 & & & Hard sigmoid & 0.24 & 0.018 \\
\br
\end{tabular}\\
\caption{\label{table:1}Table of the activation functions studied in this paper and their associated forecast horizons. The FHs shown are an average over 50 trials with reservoirs of size $N = 300$. The non-monotonic functions tend to have longer forecast horizons than the monotonic functions, but there are some exceptions. The definitions of the activation functions are in \ref{appendixA}.}
\end{table*}
In order for the RC to be able to learn nonlinear relationships with high fidelity, the activation function must be nonlinear.  In feed-forward neural networks, the activation function should vary significantly over the range of inputs it receives to avoid the ``vanishing gradient'' problem \cite{impact} during back-propagation. However, since a RC only trains the output layer weights, a vanishing gradient in the activation function may not be ruled out, and therefore there is more freedom in the choice of suitable activation functions. Historically, neural networks have used monotonic functions such as the rectified linear, logistic, Heaviside, or hyperbolic tangent functions. However, recent work has shown the success of non-monotonic activation functions \cite{searching, benchmark} for feed-forward neural networks, and Refs.~\cite{shapes, imp_ESN} have also shown that non-monotonic activation functions can perform better than monotonic ones in ESN reservoir computers.

To explore the effect of the choice of activation function on the performance of RC, we compare the FH for a collection of 16 monotonic and non-monotonic activation functions,  which are plotted in Fig.~\ref{graphs}. Some of these functions (such as the hyperbolic tangent or sigmoid) have been extensively used, while others (such as the {\it Pflu} and {\it mish}) have been only recently proposed \cite{mish,pflu}. The definitions of the functions can be found in \ref{appendixA}. For each activation function, we calculated the FH and averaged it over 50 simulations, in each of which the Lorenz training and prediction data and the random sparse reservoir were generated afresh. The averaged FHs are shown in Table~\ref{table:1}. As seen in the table, the non-monotonic functions (left) generally have a longer FH than the monotonic functions (right). However, some monotonic activation functions, such as the shifted hyperbolic tangent, the softplus, and a monotonic version of swish (see \ref{appendixA} for the definition) have high FHs. 

Table \ref{table:1} shows that the performance of a RC can be very sensitive to the activation function used. For example, the Leaky Relu and the Pflu activation functions are very close to each other (see Figure 3), but the former results in a FH approximately 3 times smaller than the latter. In addition, the functions that have the worst performance, such as the hard sigmoid, the Leaky Relu, and the R Relu, have at least one linear section. In Sec.~\ref{metrics} we will explore how the FH correlates with the curvature of the activation function.

In addition to studying the set of functions shown in Table~\ref{table:1}, we now analyze how the FH varies as we modify two specific functions, swish and shifted tanh (defined below), by changing their parameters. This allows us to see how the FH varies as the functions deform in a continuous way.

\subsection{Tuning swish}

\begin{figure}[!b]
  \begin{center}
  \includegraphics[width=0.9\linewidth]{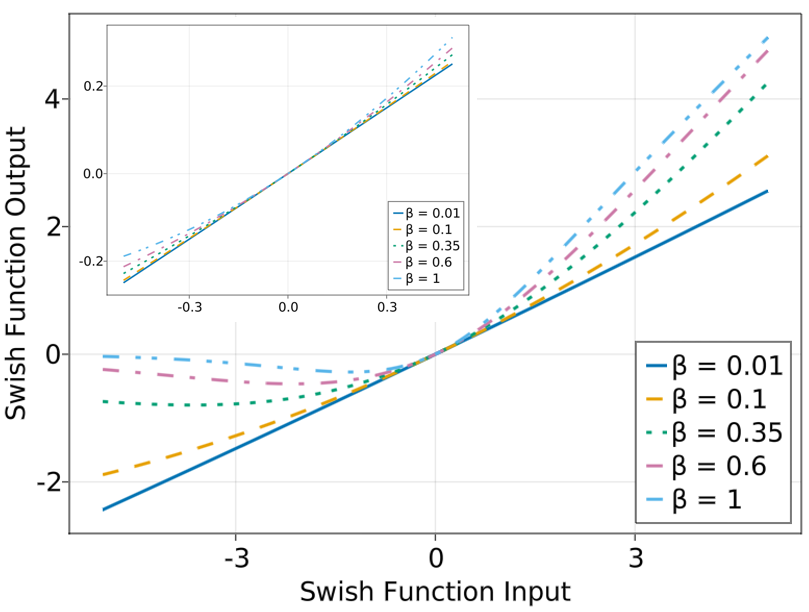}\\
  \caption{Swish function [Equation~(\ref{swish})] for $\beta = 0.01, 0.1, 0.35,0.6$, and $1$ (top to bottom). The inset is a zoomed-in view of the region $-0.5<x < 0.5$.}\label{swishes}
  \end{center}
\end{figure}

We first investigate a high-performing activation function, swish. The  swish function is defined as \cite{searching}
\begin{equation}\label{swish}
    f(x) = \frac{x}{1+e^{-\beta x}},
\end{equation}
with $\beta > 0$. 

Figure \ref{swishes} illustrates the swish function for $\beta = 0.01, 0.1, 0.35,0.6$, and $1$ (top to bottom). For $\beta = 0$ the swish function is the linear function $f(x) = x$, while in the limit $\beta \to \infty$ the swish function approaches a rectified linear unit (ReLu) function, $f(x) = \max(x,0)$. For finite and positive values of $\beta$, the swish function has a single minimum at a negative value of $x$, and it approaches $0$ and $x$ as $x$ tends to $-\infty$ and $\infty$, respectively. 

\begin{figure}[!t]
  \begin{center}
  \includegraphics[width=\linewidth]{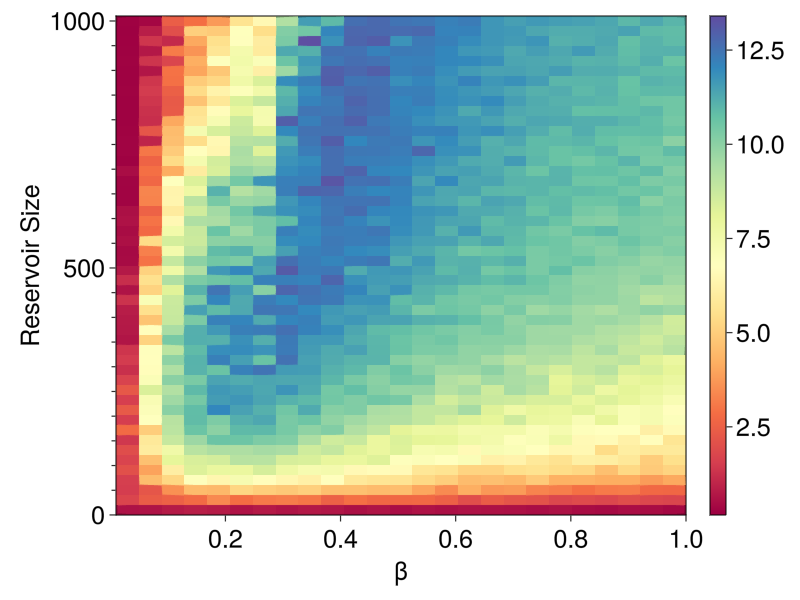}\\
  \caption{FH obtained using the swish function [Equation \eqref{swish}] as a function of the parameter $\beta$ and the reservoir size $N$. The FH was averaged over 50 trials.}\label{map_swish}
  \end{center}
\end{figure}

In order to determine how the FH varies as the swish function is varied across these extremes, we calculate the FH for 50 values of $\beta$ ranging from $0.01$ to $1$ and 50 values of $N$ ranging from 10 to 1000. For each pair $(\beta, N)$, the reservoir and training data are generated 50 different times and the resulting FHs are averaged. The resulting averaged FH is shown in Fig.~\ref{map_swish}. For small values of $\beta$ (i.e., an almost linear activation function) the FH is very low. Likewise, for small reservoirs (N $<$ 100), the FH is very low. The graph shows that for a given N, there is an optimum value of $\beta$ that yields the highest FH. 
Lastly, we note that for a given $\beta$, there is an optimal reservoir size $N$  to optimize FH; in other words, beyond a certain value of $N$, FH starts to decrease with increasing $N$. We believe this is due to limited training time and overfitting \cite{mantas}. Indeed, we are able to shift the peak FH to larger $N$ by increasing the training time and by changing the ridge regression parameter $\lambda$  (see Supplementary Material Figures S3 and S4). Thus, the graph in Figure \ref{map_swish} shows how $N$ and $\beta$ can be tuned to optimize the FH with a limited training time $T$ and a fixed ridge regression parameter $\lambda$.
\begin{figure}[!b]
  \begin{center}
  \includegraphics[width=\linewidth]{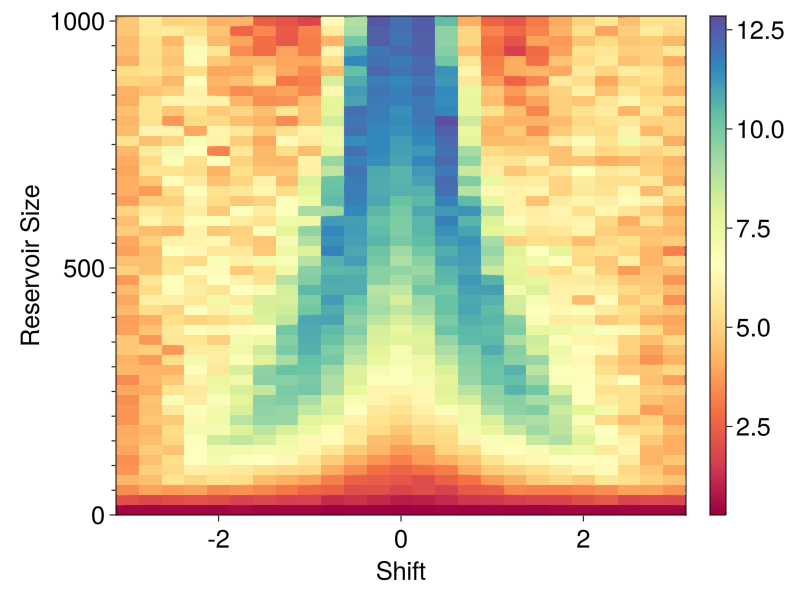}\\
  \caption{FH of shifted\_tanh (i.e. tanh(x+b)) for 25 values of b from -3 to 3 and 50 values of N ranging from 10 to 1000, averaged over 50 trials.}\label{map_tanh}
  \end{center}
\end{figure} 
\subsection{Tuning hyperbolic tangent with an input bias}
We next investigate tuning a commonly used activation function, the hyperbolic tangent with an input bias term (or shift), defined as $\text{shifted\_}\tanh(x, b) = \tanh(x+b)$. It is known that a nonzero bias term $b$ can improve the performance of reservoir computers \cite{systematic}, in some cases by making the activation function asymmetric about $0$ \cite{thesis,breaking_sym}.  We perform the same experiment as we did with swish in the previous section: we average the FH over 50 trials for 50 values of $N$ from $10$ to $1000$ and 25 values of $b$ from $-3$ to $3$.  As with swish, the training and the prediction was performed 50 times and averaged. As shown in Fig.~\ref{map_tanh}, using a bias term, especially when using small reservoirs, can significantly increase the FH. For large reservoirs, small or zero bias terms give similarly good performance for this system. This is in agreement with the results of \cite{systematic}, who find that a bias term does not significantly affect the FH for the Lorenz system (see their Fig.~6) when the reservoir is not small. These results show how the optimization of the activation function depends on the reservoir size. We note that caution should be taken when generalizing these results, since Ref.~\cite{systematic} also finds that a bias term can make a large difference when predicting the dynamics of other systems.

We also observe that for a given nonzero bias $b$, the FH reaches a maximum and eventually decreases as $N$ is increased. As with swish, changing the regularization parameter and increasing the training time causes the maximum FH to occur at larger $N$, suggesting this is due to overfitting and limited training data.

\section{Possible metrics for activation function performance}\label{metrics}

We have found that changing the activation function can greatly improve the performance of a reservoir computer. This raises the questions (i) what features of the activation function result in good performance? and (ii) what is the mechanism by which the performance is enhanced? In order to explore these questions, we study two metrics that have been hypothesized to correlate with the performance of reservoir computers. The first measure is the curvature of the activation function (a proper definition will be given below). We also study the entropy of reservoir internal states, which has been previously been suggested to correlate with the reservoir's performance. 

\begin{figure}[t]
  \begin{center}
  \includegraphics[width=\linewidth]{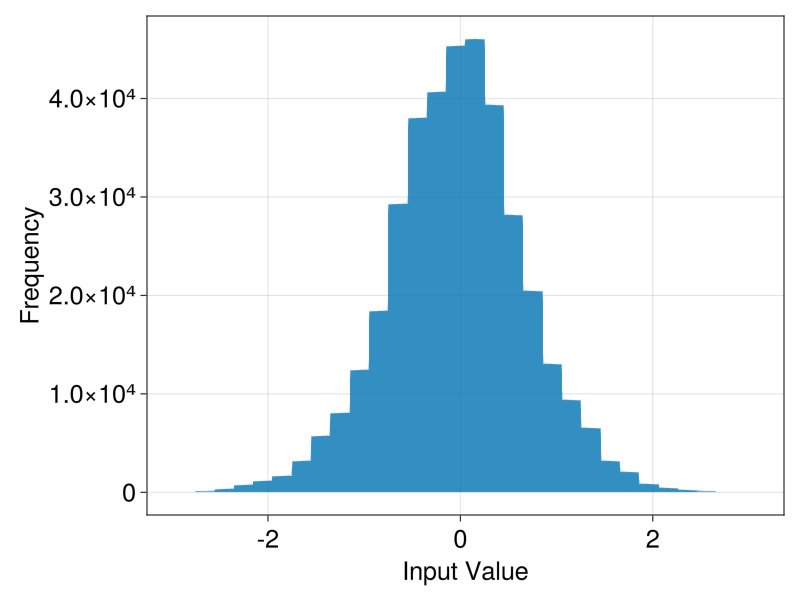}\\
  \caption{Histogram of input values to the tanh activation function, for N = 300.}\label{hist_tanh}
  \end{center}
\end{figure} 

\begin{figure}[!h]
\centering
\text{(A)}
\includegraphics[width=0.75\linewidth]{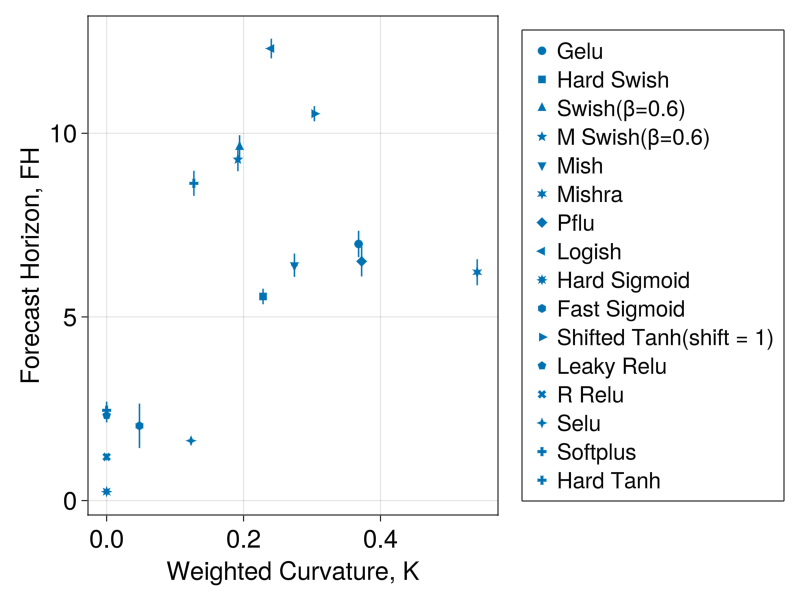}\\
\text{(B)}
\includegraphics[width=0.75\linewidth]{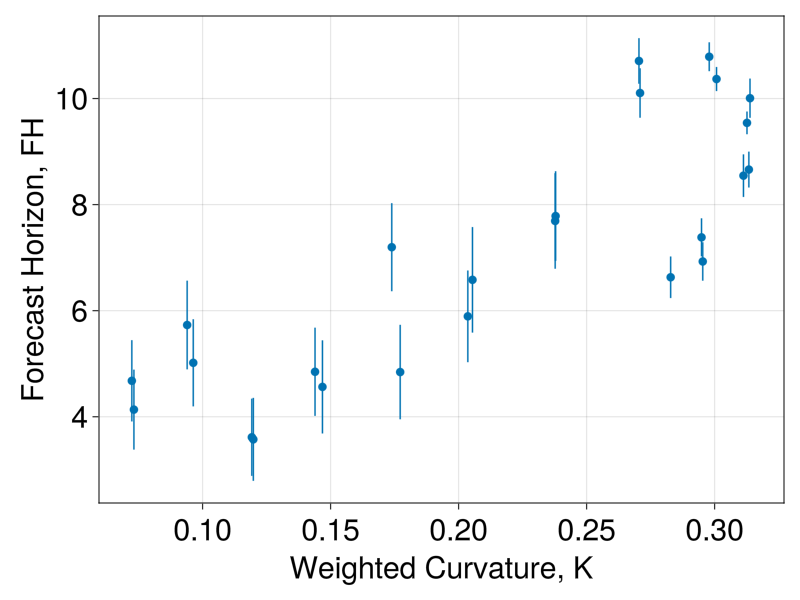}\\
\text{(C)}
\includegraphics[width=0.75\linewidth]{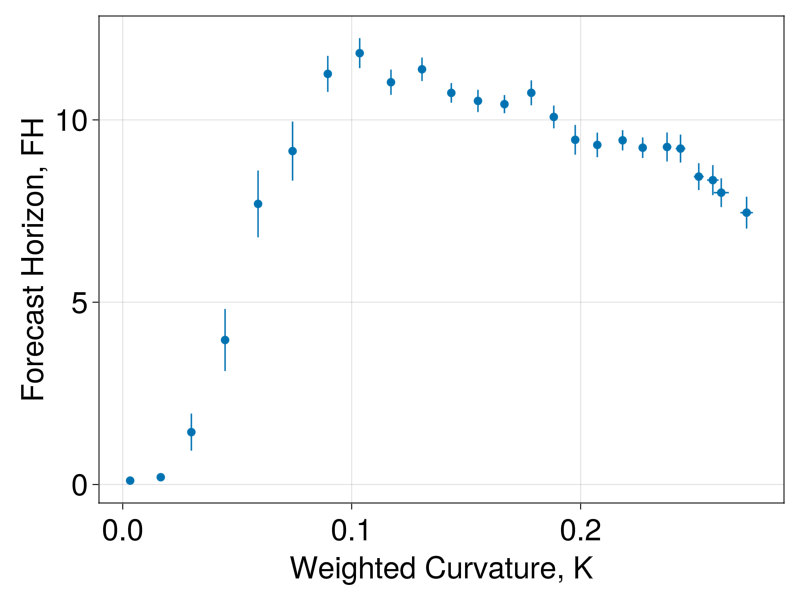}\\
\text{(D)}
\includegraphics[width=0.75\linewidth]{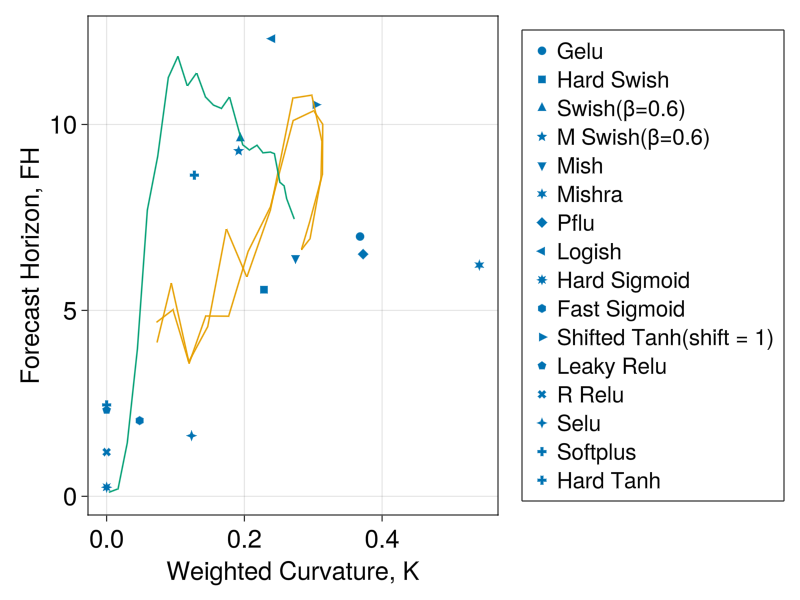}
\caption{FH versus $K$ for (A) the set of 16 functions on Table~\ref{table:1}, (B) the swish function with 25 different values of $\beta$, (C) the $\text{shifted\_}\tanh$ function with 25 different values of $b$, and (D) all previous cases on the same figure. In (A), (B), and (C), the vertical error bars show the standard error of the FH calculated over the $50$ realizations, and the horizontal error bars show the standard error of $K$ calculated over the $50$ realizations (in some cases, the horizontal error bars are too small to see). In (D), the swish and $\text{shifted\_}\tanh$ points are connected with yellow and green lines, respectively.}
\label{FHvsK}
\end{figure}

\subsection{Weighted average curvature}

As mentioned earlier, a common feature of activation functions with poor performance is the presence of linear portions in the function. This suggests that the degree of nonlinearity should be correlated with the performance of activation functions. Therefore, we aim to compare the FH of activation functions with a measure of their nonlinearity, a surrogate of which is the function's curvature. Since the only relevant parts of the activation function are those that are actually used during the operation of the reservoir, we define the {\it weighted curvature} $K$ of an activation function $f$ as
\begin{align}
K = \int \kappa(x) p(x) dx,
\end{align}
where 
\begin{align}
\kappa(x) = \frac{|f^{''}(x)|}{(1 + f(x)^{'2})^{3/2}}
\end{align}
is the curvature of the activation function at a point $x$, and $p(x)$ is the probability that the activation function is evaluated at $x$ during the operation of the reservoir. In practice, this probability is estimated numerically by binning the relevant range of activation function inputs and constructing a histogram of these inputs during the operation of the reservoir. Such a histogram for the hyperbolic tangent is shown in Figure \ref{hist_tanh}; histograms for  the other activation functions can be found in the Supplementary Material. Note that, because $p(x)$ depends on the input and parameters of the reservoir, the weighted average curvature $K$ depends not only on the activation function $f$, but also on the reservoir parameters and the system to be studied (in our case, the Lorenz system). 

Figure~\ref{FHvsK} shows FH versus $K$ for (A) the set of 16 functions on Table~\ref{table:1}, (B) the swish function with 25 different values of $\beta$, (C) the $\text{shifted\_}\tanh$ function with 25 different values of $b$, and (D) all previous cases on the same figure. In (A), (B), and (C), the vertical error bars show the standard error of the FH calculated over the $50$ realizations, and the horizontal error bars show the standard error of $K$ calculated over the $50$ realizations (in some cases, the horizontal error bars are too small to see). In (D), the swish function points are connected with a yellow line to aid the eye, while the $\text{shifted\_}\tanh$ points are connected with a green line. 

Overall, Fig.~\ref{FHvsK}(D) shows there is a positive correlation of FH with weighted curvature $K$ for the activation functions studied. In particular, all the activation functions with low weighted curvature (less than approximately $0.05$) have a small forecast horizon. However, some activation functions with high weighted curvature (e.g., mishra) do not have particularly high FH. In addition, while Fig.~\ref{FHvsK}(B) shows a robust positive correlation between FH and $K$ for the swish function, Fig.~\ref{FHvsK}(C) shows that for the $\text{shifted\_}\tanh$ function the FH is maximized at an intermediate value of weighted curvature, and decreases slightly beyond this value.

In summary, we found that there is a positive correlation between weighted curvature $K$ and FH. In particular, very small weighted curvature results in low FH. Howewer, weighted curvature alone is not a good predictor of FH for an individual activation function. 

\begin{figure}[h!]
\centering
\text{(A)}
\includegraphics[width=0.75\linewidth]{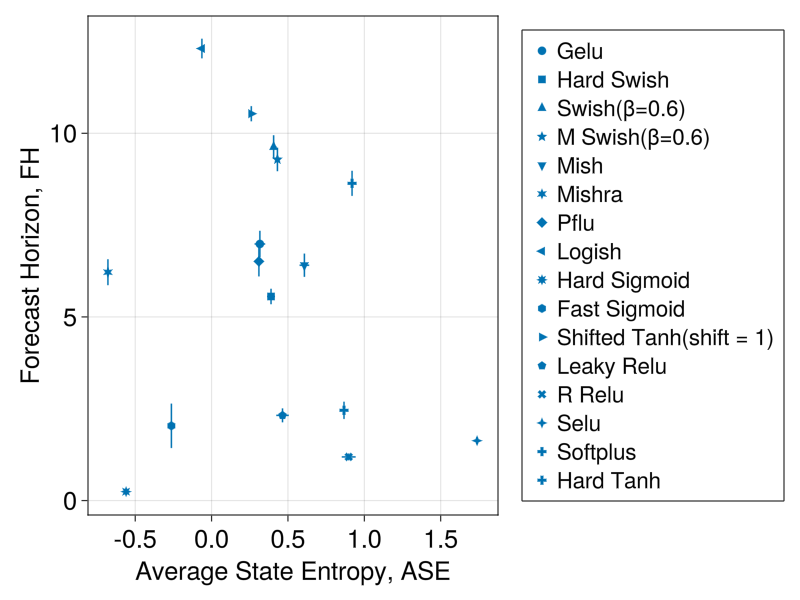}\\
\text{(B)}
\includegraphics[width=0.75\linewidth]{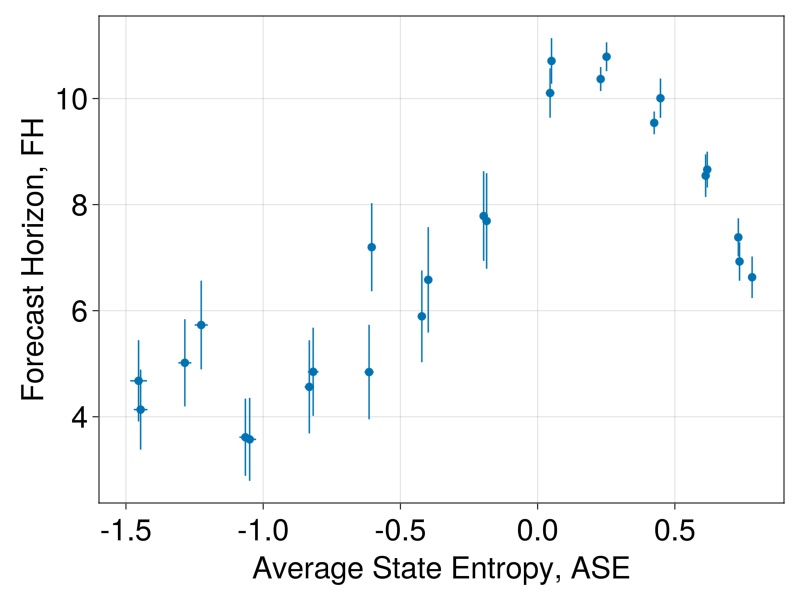}\\
\text{(C)}
\includegraphics[width=0.75\linewidth]{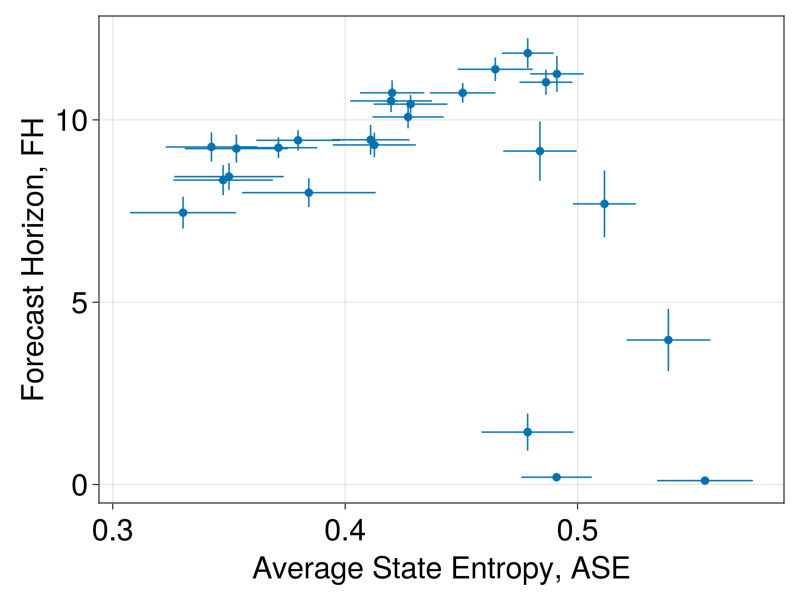}\\
\text{(D)}
\includegraphics[width=0.75\linewidth]{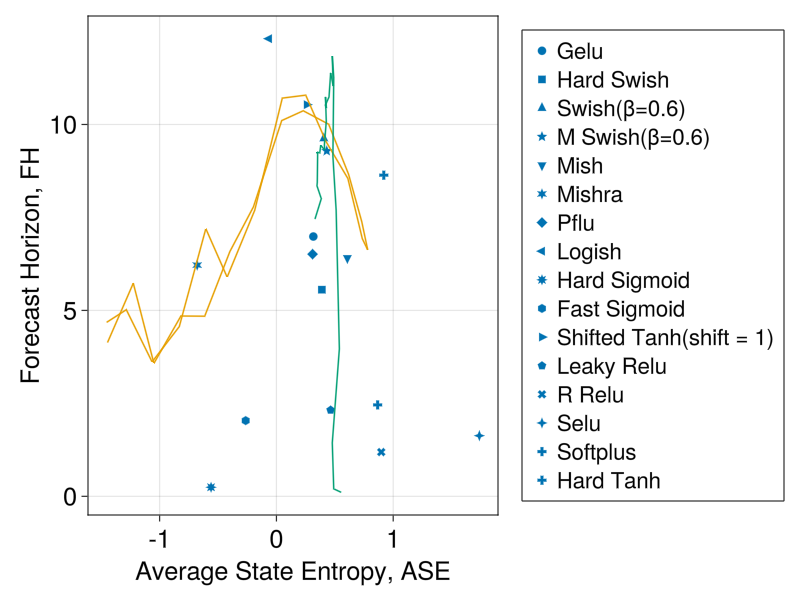}
\caption{FH versus ASE for (A) the set of 16 functions on Table~\ref{table:1}, (B) the swish function with 25 different values of $\beta$, (C) the $\text{shifted\_}\tanh$ function with 25 different values of $b$, and (D) all previous cases on the same figure. In (A), (B), and (C), the vertical error bars show the standard error of the FH calculated over the $50$ realizations, and the horizontal error bars show the standard error of ASE calculated over the $50$ realizations. In (D), the swish and $\text{shifted\_}\tanh$ points are connected with yellow and green lines, respectively.}\label{FHvsASE}
\end{figure}

\clearpage
\subsection{Average State Entropy (ASE)}

Another quantity that has been proposed to correlate with the performance of reservoir computers is the entropy of the internal reservoir states \cite{ase1,entropy1}. Intuitively, a higher entropy corresponds to richer dynamics, and thus to the possibility of encoding more complex behavior in the reservoir. Since directly calculating the entropy of states would be computationally impractical, we follow Refs.~\cite{ase1,entropy1} and use an estimator of Renyi's quadratic entropy, the instantaneous state entropy (ISE), defined as
\begin{equation}
    \text{ISE}(t) = - \log \left[ \frac{1}{N^2} \sum_j  \sum_i K_\sigma (r_j(t) - r_i(t)) \right],
\end{equation}
where ${\bf r}(t)$ is the internal state of the reservoir at time $t$, $K_{\sigma}(x) = \exp(-x^2/(2\sigma^2))/(\sqrt{2\pi}\sigma)$ is a Gaussian function with standard deviation $\sigma$, and $\sigma$ is taken to be $0.3$ times the standard deviation of the internal state vector ${\bf r}$ \cite{ase1}. We define the {\it average state entropy} ASE as the time average of the ISE from the first time step of prediction up to the FH.  

Figure~\ref{FHvsASE} shows FH versus ASE for (A) the set of 16 functions on Table~\ref{table:1}, (B) the swish function with 25 different values of $\beta$, (C) the $\text{shifted\_}\tanh$ function with 25 different values of $b$, and (D) all previous cases on the same figure. In (A), (B), and (C), the vertical error bars show the standard error of the FH calculated over the $50$ realizations, and the horizontal error bars show the standard error of the ASE calculated over the $50$ realizations. As in Fig. \ref{FHvsK}(D), the swish function points in (D) are connected with a yellow line, and the $\text{shifted\_}\tanh$ points are connected with a green line. 

The dependence of FH on ASE for the collection of functions shown in (A) is not clear. For the swish and shifted\_tanh functions, for which a more definite pattern is observed, the maximum FH occurs at intermediate values of the ASE. This is consistent with the findings of Ref.~\cite{carroll_entropy2}. In this reference, the performance of reservoir computers with polynomial activation functions was studied. It was found that although the entropy of the reservoir internal states increases as the edge of stability of the reservoir is approached (i.e., the point at which the reservoir dynamics becomes unstable) the best performance is not always achieved at this point. In agreement with these observations, we see in Figs.~\ref{FHvsASE}(B) and (C) that the best reservoir performance does not occur at the point of maximum entropy. 

\section{Discussion}\label{conclusion}

In this work, we investigated how the activation function of a RC can be tuned to increase the FH in predicting the evolution of the chaotic Lorenz system. We found that under fixed conditions, changing the activation function alone can have a dramatic effect on the FH, with FH values ranging from nearly zero to nearly thirteen Lyapunov times for the activation functions we studied. While we found that non-monotonic activation functions generally performed better than monotonic ones, some monotonic functions performance was comparable to that of non-monotonic functions.

The wide variety in performance for activation functions that, on the surface, do not appear so different (see Figure~\ref{graphs}), led us to investigate what factors affect the performance of the reservoir under the different activation functions. We explored the dependence of the FH obtained using different activation functions on the curvature of the activation function and on the entropy of the internal reservoir states. We studied this dependence both for the set of 16 different activation functions in Figure~\ref{graphs}, and additionally for two particular activation functions modified by changing a parameter (swish and shifted hyperbolic tangent). We found that there is a positive (although not strong) correlation between the FH and the weighted curvature of the activation function used. In particular, activation functions that result in small weighted curvature yield poor performance. However, as seen in Figure~\ref{FHvsK}(C), weighted curvature does not predict FH. The dependence of FH on the internal state entropy, which we quantified using the Average State Entropy ASE, was less clear. In agreement with previous results \cite{carroll_entropy2}, we found that the point of maximum performance (i.e., maximum FH) does not coincide with the point of maximum entropy for the task studied in this paper. Finding a way to predict which activation function will result in a better performance for a given task remains an elusive goal.

We also studied systematically how the performance of the FH changes when the activation function is changed continuously by modifying a parameter (Figures~\ref{map_swish} and \ref{map_tanh}). For fixed reservoir parameters, modifying the activation function can result in tremendous improvements in performance: for example, note how there is a dramatic improvement in performance for $N = 1000$ as $\beta$ is increased from $\beta \sim 0.1$ to $\beta \sim 0.4$. 

We now discuss the important issue of hyperparameter optimization. In this paper, we studied how the performance of the RC changes when the activation function is changed, keeping the rest of the parameters constant. One might perform a hyperparameter optimization for each activation function used, and obtain different (likely better) values of the FH for each \cite{bergstra}. Such a procedure would improve the performance of each activation function and might change the relative performance among activation functions, since the fixed set of parameters we used might favor some activation functions over others. However, in this paper we wanted to illustrate how  the performance of the RC can be increased by just changing the activation function without doing an exhaustive hyperparameter optimization. We acknowledge that if further performance improvements are sought, other parameters need to be optimized as well.

One limitation of our work is that it is restricted to exploring the performance of a fixed and somewhat arbitrary set of prescribed activation functions. In principle, one could seek to construct more exotic, better performing activation functions either by optimization principles such as genetic algorithms or by other mechanisms (as was done in Refs.~\cite{searching, genetic} for feed-forward neural networks). Our work provides a strong motivation for this endeavor. Another limitation is that we restricted our study to the Lorenz system. While the performance of specific activation functions will surely be different for other systems, we believe that the fact that there is plenty of room for performance improvement by tuning the activation function is true for other systems.

In conclusion, our results show that the performance of a reservoir computer can be significantly improved by modifying the activation function. Hyperparameter optimization algorithms do not usually modify the activation function, but focus instead on other reservoir parameters. The large variation in performance among the activation functions considered here suggests that including some variation in the activation function during hyperparameter optimization algorithms might lead to improvement in the performance of reservoir computers.



\section*{Acknowledgments}

The authors would like to thank the University of Colorado Boulder College of Engineering and Applied Science Interdisciplinary Research Theme (IRT) program and the Renewable and Sustainable Energy Institute (RASEI) for seed grant funds to support this work.

\appendix

\section{Activation Functions}\label{appendixA}

Below are the definitions for the activation functions in Table 1 in the main text. Graphs of the activation functions are shown in Figure \ref{graphs}. In the definitions below, $\sigma(x) = 1/(1+\exp(-x))$ and $\Phi(x)$ is the standard Gaussian cumulative distribution function. We used the library NNlib.jl to calculate many of the functions.

\begin{table}[h]
  \centering
  \begin{tabular}{p{0.3\linewidth} p{0.65\linewidth}}
  \br
{\bf Name} & {\bf Definition}\\
\mr
Gelu & $x \cdot \Phi(x)$\\
\hline
Hard Swish & $x \cdot \max(0, \min(1, (x+3)/6))$\\
\hline
Logish & $x\cdot \ln(1+ \sigma(x))$ \cite{logish}\\
\hline
Mish & $x \cdot \tanh (\text{softplus}(x))$ \cite{mish}\\
\hline
Mishra & $\frac{1}{2}\left( \frac{x}{1+ |x|}\right)^2 + \frac{1}{2}\frac{x}{1+ |x|}$ \cite{benchmark}\\
\hline
Pflu & $x \cdot \frac{1}{2} \left( 1 + \frac{x}{\sqrt{1+x^2} + \frac{x}{(1+x^2)\sqrt{1+x^2}}} \right)$ \cite{pflu}\\
\hline
Swish & $x \cdot \sigma(x)$\\
\hline

  \end{tabular}
    \caption{Non-monotonic activation functions}
\end{table}

\begin{table}[h]
  \centering
  \begin{tabular}{p{0.3\linewidth} p{0.65\linewidth}}
  \br
{\bf Name}& {\bf Definition}\\
\mr
Sigmoid & $\sigma(x)=\frac{1}{1+e^{-x}}$\\\\
\hline
Shifted$\_$tanh & tanh(x + b)\\
\hline
Hard Sigmoid& $\max(0, \min(1, (x+3)/6))$\\
\hline
Hard Tanh & $\max(-1, \min(1,x))$\\
\hline
Leaky Relu & $\max(ax,x)$, $a=0.01$\\
\hline
Monotonic & $ \text{swish}(x)$ if $ x > x_{\text{min}}$ \\
Swish  &  $ \text{swish}(x_\text{min})$ if $x \leq x_{\text{min}}$\\
&$x_{\text{min}} = \text{argmin}(\text{swish}(x))$\\
\hline
R Relu & $\max(ax,x)$\\
& For each $x$, $a$ is randomly sampled from a uniform distribution in (1/8, 1/3)\\
\hline
Selu & $\lambda x$ if $x >0$\\
& $\lambda(\alpha e^x - \alpha)$ if $x \leq 0$\\
& $\lambda = 1.05070$, $\alpha = 1.67326$\\
\hline
Softplus &  $\log(e^x + 1)$ \\
\hline
  \end{tabular}
    \caption{Monotonic activation functions}

\end{table}
With some activation functions, we encountered a problem where the inputs increased indefinitely as the reservoir operation repeatedly iterates the activation function. To prevent this, we introduced a bound $B$ on all the activation functions considered by modifying them as
\begin{eqnarray}\label{biswish}
    \text{function}(x) \to \left\{
    \begin{array}{lc}
    \text{function}(x), &-B < x < B\\
    \text{function}(B), &x \geq B\\
    \text{function}(-B), &x \leq -B
\end{array}
\right.
\end{eqnarray}
where the bound B is an adjustable parameter which we set to $B= 5$ (See SM).

\bibliographystyle{unsrt}

\end{document}


\title[Tuning the activation function to optimize the forecast horizon of a RC]{Supplementary material: Tuning the activation function to optimize the forecast horizon of a reservoir computer}

\author{L A Hurley$^1$, J G Restrepo$^2$ and S E Shaheen$^{1,3,4}$}

\address{$^1$Department of Electrical, Computer and Energy Engineering, University of Colorado at Boulder}
\address{$^2$Department of Applied Mathematics, University of Colorado at Boulder}
\address{$^3$Renewable and Sustainable Energy Institute (RASEI)}
\address{$^4$Department of Physics, University of Colorado at Boulder}
\ead{lauren.hurley@colorado.edu}
\vspace{10pt}
\begin{indented}
\item[]December 2023
\end{indented}




%
%
%
%
%

\section{Simulation algorithm}
This is a description of the algorithm used in the Echo State Network (ESN) simulation, with the Julia package ReservoirComputing.jl. For more information, please refer to the documentation for that package. For each trial,

\begin{enumerate}
    \item The Lorenz data is generated.
    \item The data is split into two sets, training and prediction. We used 5000 time units for the training and 1250 time units for prediction.
    \item The reservoir is generated according to the parameters specified in the main text.
    \item The output layer is trained using the training data and the specified training method as discussed in the main text.
    \item The FH is calculated as the first instance where the output differs from the testing data by more than 5 (since 5 is about 15\% of the overall output of the x-coordinate).
\end{enumerate}

At the end of all the trials, the FH value is averaged.

\clearpage
\section{Curvature of activation functions}
\begin{figure}[h]
  \begin{center}
  \includegraphics[width=\textwidth]{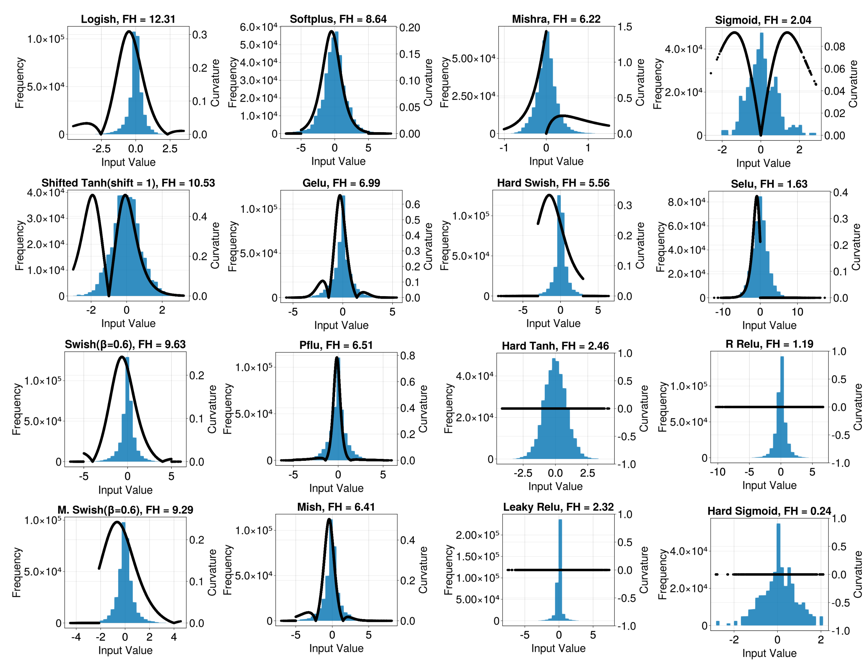}\\
  \caption{Graphs of curvature (solid black lines) and histograms (blue bars) of input values for the activation functions discussed in the main text; each title states the name of the function and the average FH for 50 trials in a 300-node reservoir.}\label{curvs}
  \end{center}
\end{figure}

In Figure \ref{curvs} we show the curvature (solid black line) and a histogram of the inputs (blue bars) for the 16 activation function discussed in the main text. \\

\clearpage
\section{Increasing training time}

\begin{figure}[h]
  \begin{center}
  \includegraphics[width=6in]{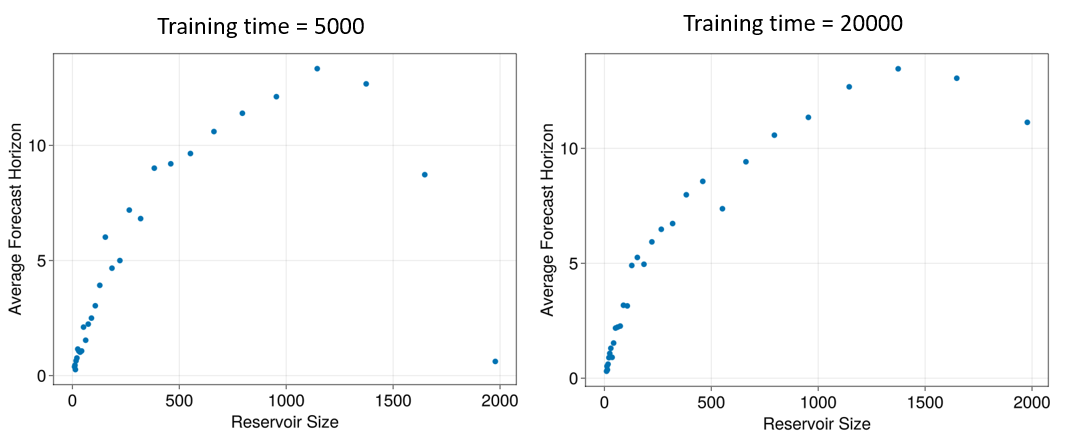}\\
  \caption{Average FH (10 trials) versus $N$ for the tanh activation function with 5000 (left) and 20000 (right) training time. Increasing the training time causes larger reservoirs to have higher FH.}\label{tanh}
  \end{center}
\end{figure}

\begin{figure}[h]
  \begin{center}
  \includegraphics[width=6in]{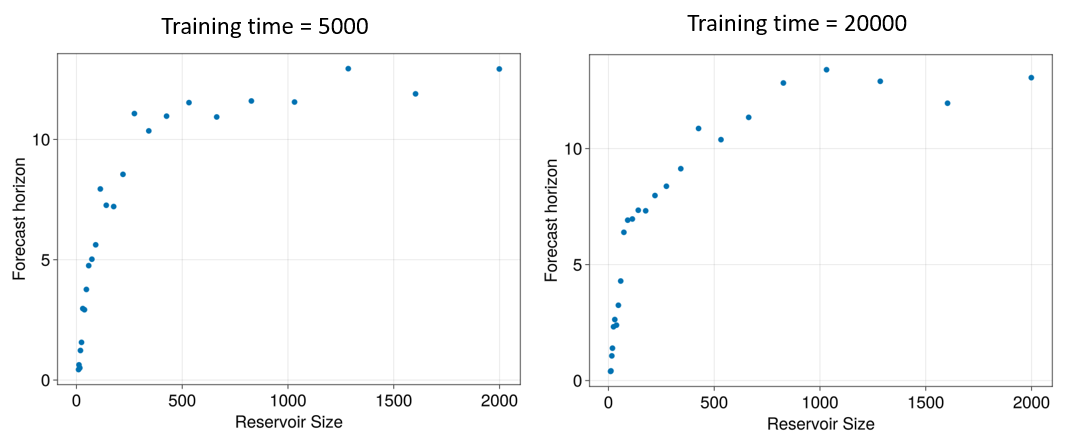}\\
  \caption{Average FH (10 trials) vs $N$ for swish with parameters $\beta$=0.6 and  bound $B = 5$ with 5000 (left) and 20000 (right) training time. Increasing the training time causes larger reservoirs to have higher FH.}\label{biswish}
  \end{center}
\end{figure}

Increasing the training time tends to cause the maximum FH to occur at larger N. Figures \ref{tanh} and \ref{biswish} illustrate this behavior for both tanh and swish.

\clearpage
\section{Optimizing Ridge Regression Parameter $\lambda$}

\begin{figure}[h]
  \begin{center}
  \includegraphics[width=6in]{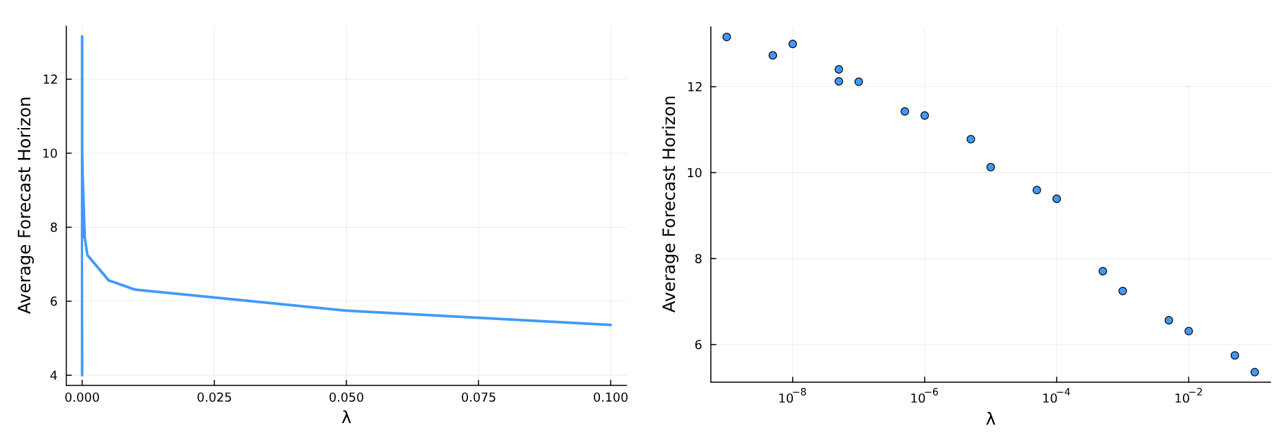}\\
  \caption{Average FH (50 trials) versus $\lambda$ for an activation function of $\tanh{(x+1.7)}$ and $N = 1000$, plot on a linear axis (left) and a logarithmic axis (right). $\lambda$ can be optimized to yield a higher FH.}\label{rr}
  \end{center}
\end{figure}

The ridge regression parameter $\lambda$ can be tuned to optimize the FH while holding all other parameters fixed. For example, in Figure \ref{rr} we choose conditions which yield a poor FH ($N = 1000$ and an activation function of $\tanh{(x+1.7)}$) and demonstrate how changing $\lambda$ can improve the FH. 

\clearpage
\section{Varying the input bound on swish}
\begin{figure}[h]
  \begin{center}
  \includegraphics[width=6.4in]{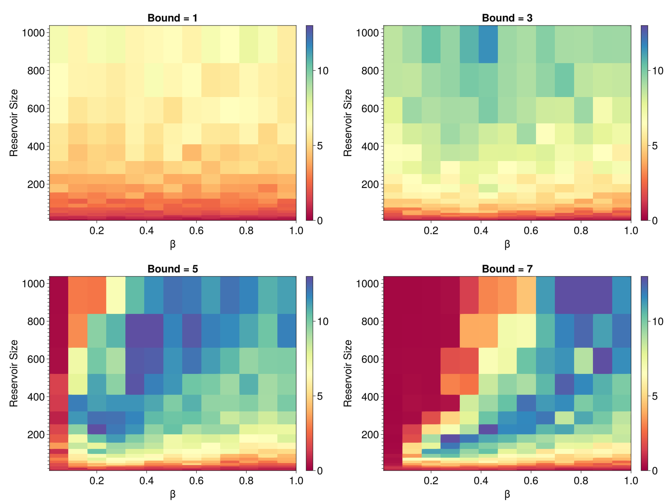}\\
  \caption{Average FH (10 trials) as a function of N and $\beta$ for swish with different input bounds $B$ (see activation function definitions in the main text). Increasing the bound initially increases the maximum FH, but after a bound of $B = 5$, increasing the bound no longer increases the FH and decreases the region of values of $\beta$ with high FH.}\label{bounds}
  \end{center}
\end{figure}

The plots in Figure \ref{bounds} show the FH as a function of N and $\beta$ for swish with different input bounds $B$ (see activation function definitions in the main text). Starting from a bound of $B=0$, increasing the bound increases the FH until around $B=5$. For bounds larger that $B=5$, the FH does not exceed 13 and the region of high-performing values of $\beta$ shrinks.

\clearpage
\section{Distribution of FH values over several trials}
\begin{figure}[h]
  \begin{center}
  \includegraphics[width=5in]{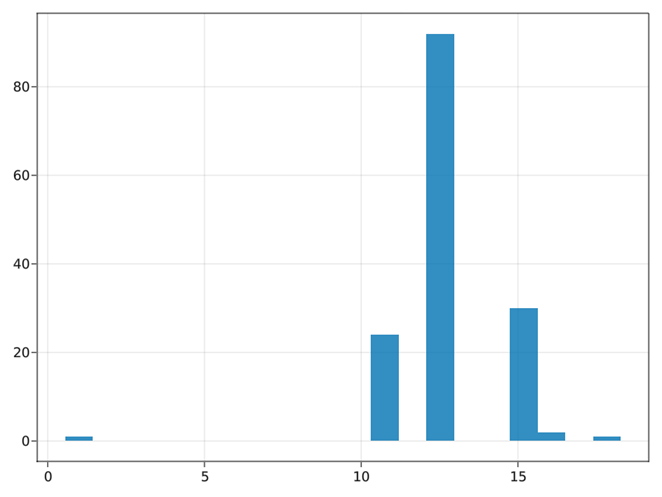}\\
  \caption{Histogram (20 bins) of FH values for 150 trials with the same activation function (swish with  $\beta = 0.45$) and the same reservoir size ($N = 800$).}\label{FH_hist}
  \end{center}
\end{figure}

Figure \ref{FH_hist} is a histogram (with 20 bins) of FH values for 150 trials with the same activation function (swish with  $\beta = 0.45$) and the same reservoir size ($N = 800$). The mean is 12.65 and the standard deviation is 1.9. We note that the distribution is somewhat discrete, which suggests that there are preferred FH values for given conditions. This is presumably due to the fact that trajectories diverge from each other at different rates in different places of the attractor, and therefore predictions are more likely to diverge from the actual trajectory at those regions.

\clearpage
\section{How ISE changes in time}
\begin{figure}[h]
  \begin{center}
  \includegraphics[width=5in]{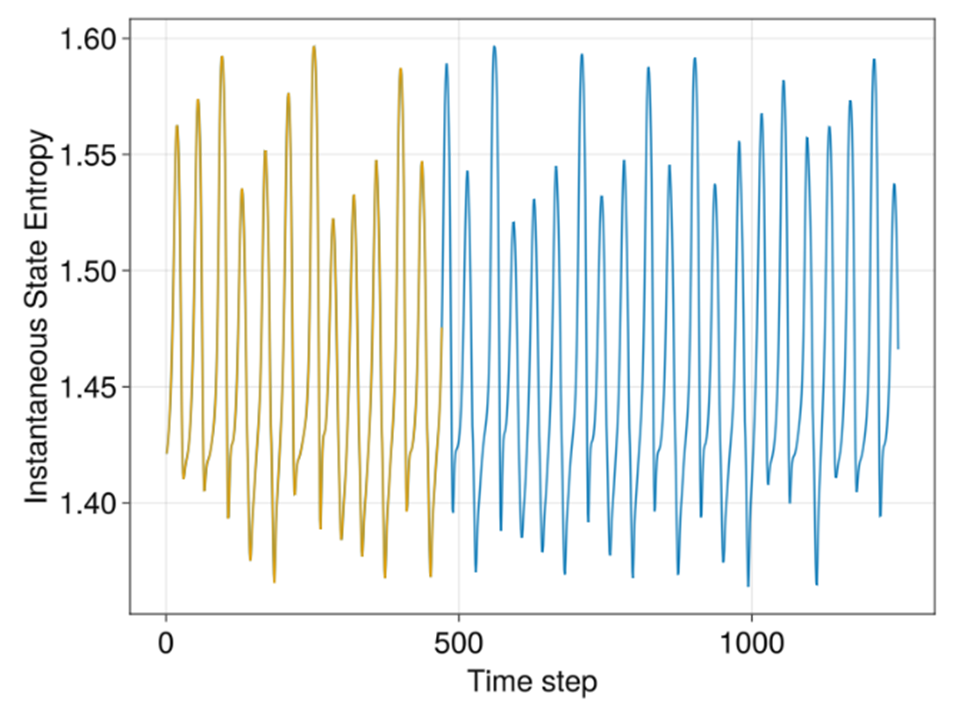}\\
  \caption{ISE versus time step for swish($\beta=0.2$) and N = 312. The ISE is plot in yellow up until the FH and in blue after the FH.}\label{ise}
  \end{center}
\end{figure}

We focused on studying the average state entropy (ASE), the time average of the instataneous state entropy (ISE). In Fig.~\ref{ise} we plot the instantaneous state entropy as a function of time for swish with bound $\beta = 0.2$ and $N = 312$.  The ISE is plot in yellow up until the FH and in blue after the FH.

